\documentclass[a4paper,conference]{IEEEtran}
\usepackage{graphicx}
\usepackage{amsmath,amssymb} 
\usepackage[linesnumbered,ruled,vlined]{algorithm2e}
\usepackage{url}
\usepackage{amsfonts}       
\usepackage{nicefrac}       
\usepackage{subfigure}
\usepackage[table]{xcolor}
\usepackage[super]{nth}
\usepackage[flushleft]{threeparttable}
\DeclareMathOperator*{\argmin}{arg\,min}

\begin{document}

\title{The Surprising Effectiveness of Linear Unsupervised Image-to-Image Translation}

\author{\IEEEauthorblockN{Eitan Richardson}
\IEEEauthorblockA{School of Computer Science and Engineering\\
The Hebrew University of Jerusalem\\
Jerusalem, Israel\\
Email: eitanrich@cs.huji.ac.il}
\and
\IEEEauthorblockN{Yair Weiss}
\IEEEauthorblockA{School of Computer Science and Engineering\\
The Hebrew University of Jerusalem\\
Jerusalem, Israel\\
Email: yweiss@cs.huji.ac.il}}

\maketitle

\begin{abstract}
Unsupervised image-to-image translation is an inherently ill-posed problem. Recent methods based on deep encoder-decoder architectures have shown impressive results, but we show that they only succeed due to a strong locality bias, and they fail to learn very simple nonlocal transformations (e.g.\ mapping upside down faces to upright faces). When the locality bias is removed, the methods are too powerful and may fail to learn simple local transformations. In this paper we introduce {\em linear} encoder-decoder architectures for unsupervised image to image translation. We show that learning is much easier and faster with these architectures and yet the results are surprisingly effective. In particular, we show a number of local problems for which the results of the linear methods are comparable to those of state-of-the-art architectures but with a fraction of the training time, and a number of nonlocal problems for which the state-of-the-art fails while linear methods succeed.
\end{abstract}

\IEEEpeerreviewmaketitle

\section{Introduction}
In unsupervised image-to-image translation we are given a set of images from domain $A$ (e.g.\ black and white images of faces) and a set of images from domain $B$ (e.g.\ color images of faces). We do not know the correspondence between images in the two sets (in fact, such a correspondence might not exist), and we nevertheless seek to learn a mapping from domain $A$ to $B$. In a probabilistic view of the same problem, there exists some joint distribution $P_{A, B}$ over the two domains. We are given iid samples from the two \emph{marginal} distributions $P_A$ and $P_B$ and we want to infer $P_{B|A}$.

\begin{figure}
    \centering
    \subfigure[]{
    \includegraphics[height=2.25cm]{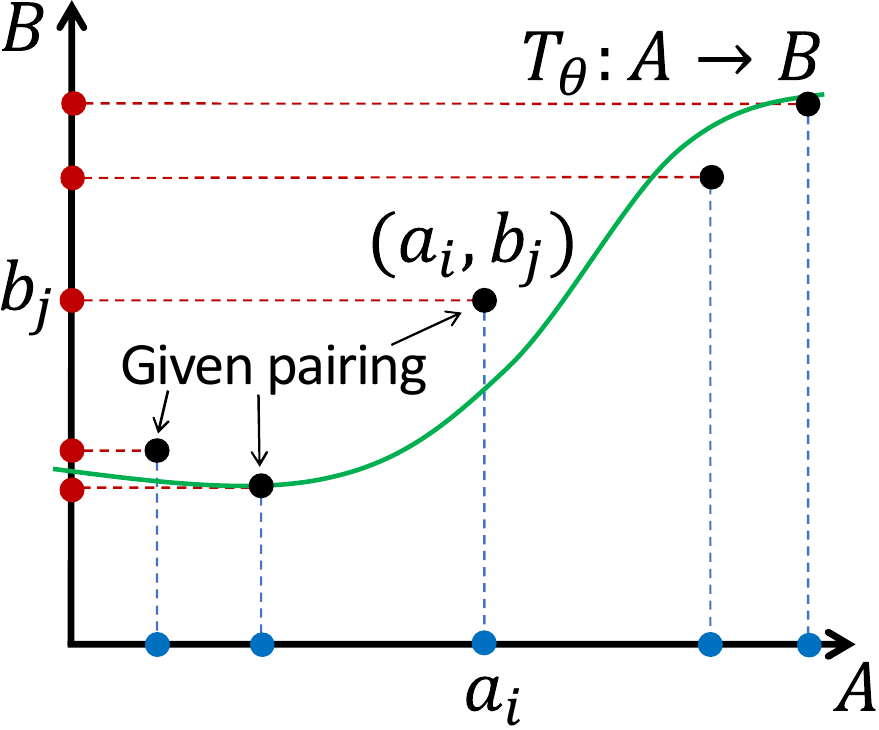}}
    \subfigure[]{
    \includegraphics[height=2.25cm]{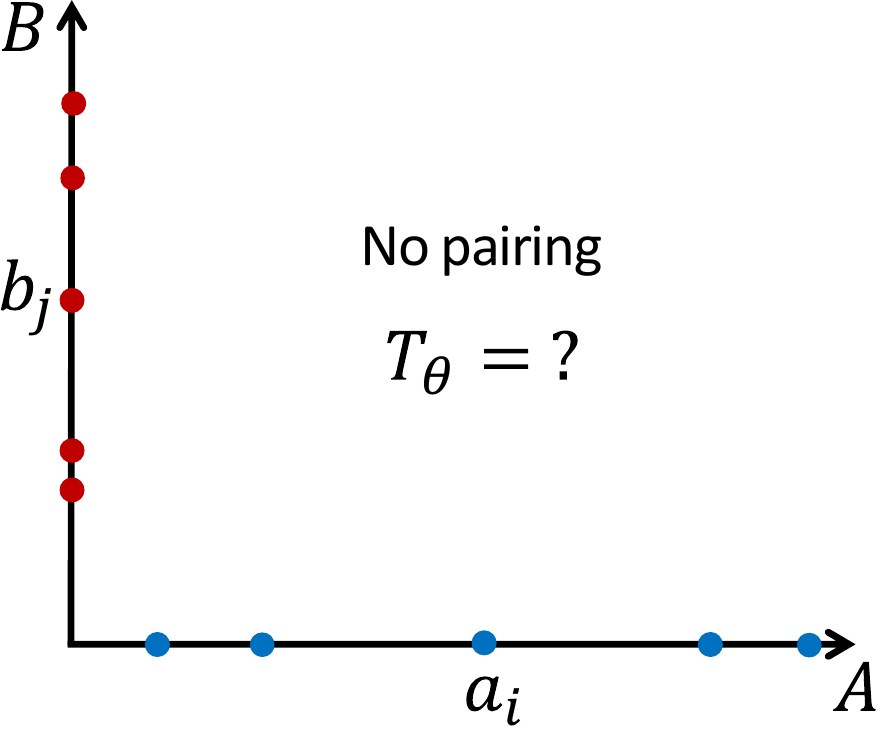}}
    \subfigure[]{
    \includegraphics[height=2.25cm]{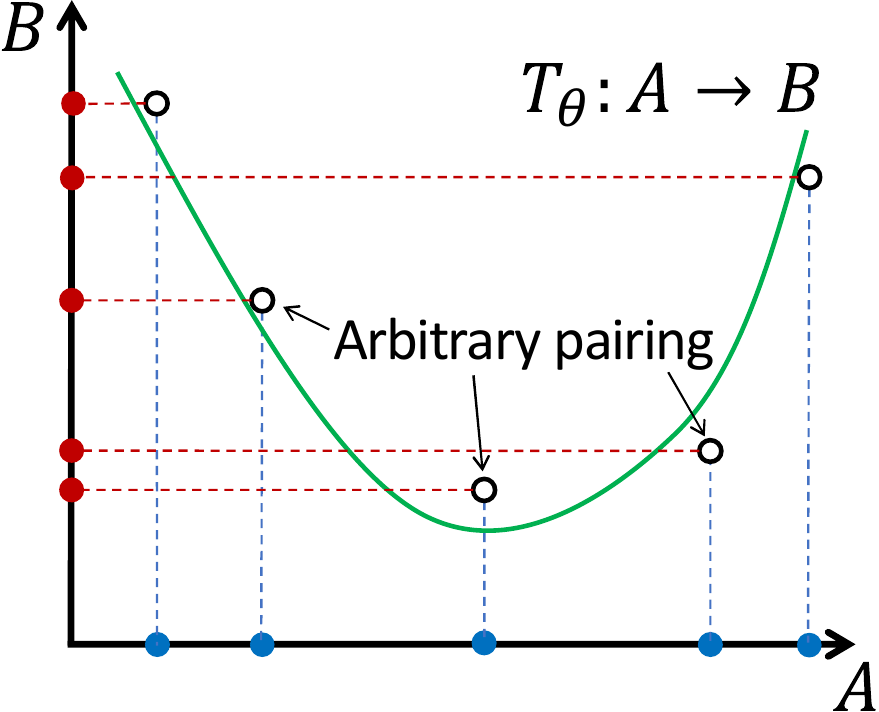}}
    \caption{(a) When pairing is given as supervision, domain translation is a regression problem -- fitting some parametric transformation $T_{\theta}$. (b) Without the pairs correspondence, the problem is ill-posed: (c) any arbitrary correspondence can be chosen, resulting in an arbitrary learned transformation.}
    \label{fig:udt-2d-schematic}
\end{figure}

This problem is inherently ill-posed. We can always define an arbitrary correspondence of the images in the two sets and learn a mapping from each image in $A$ to its corresponding image in $B$.
This is a perfectly valid mapping from $A$ to $B$. Figure~\ref{fig:udt-2d-schematic} illustrates the ill-posedeness when both $A$ and $B$ are one dimensional.

\begin{figure}
    \centering
    \includegraphics[width=0.85\columnwidth]{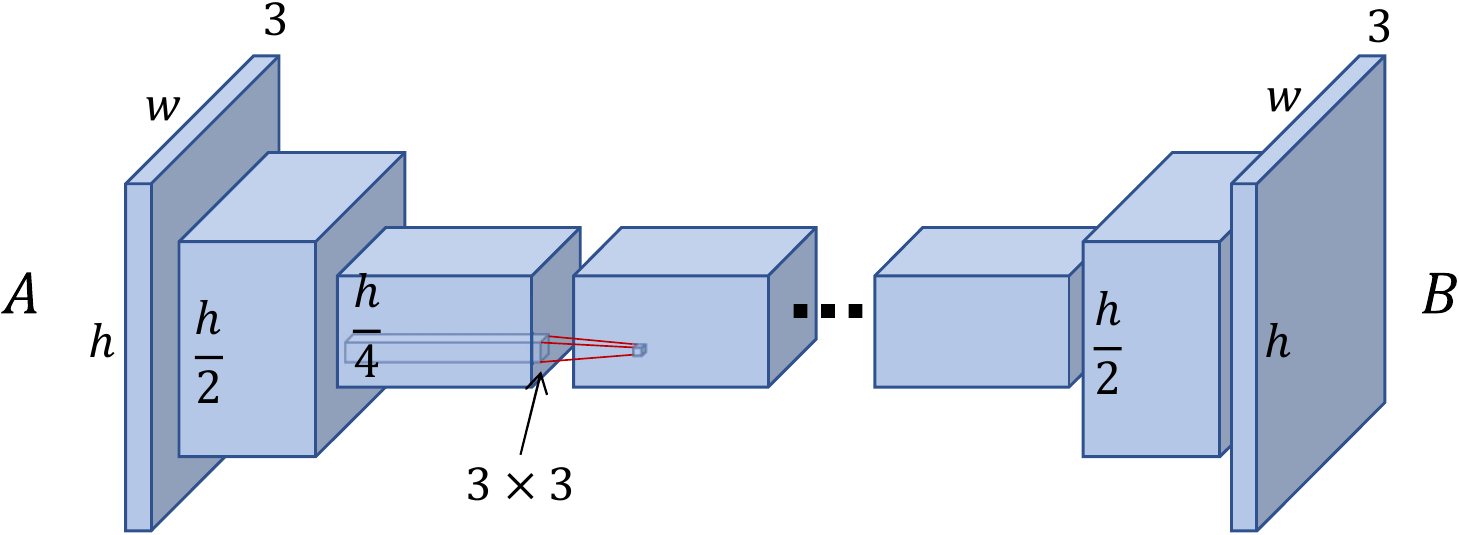}
    \caption{The strong locality bias in CycleGAN and other \emph{im2im} methods is mainly due to the large spatial dimension of the encoder-decoder bottleneck, typically $1/4$ of the input resolution, combined with a small convolution kernel size (e.g.\ $3\times3$).}
    \label{fig:bias-cyclegan}
\end{figure}

Despite this inherent ambiguity, significant progress has been achieved in recent years on this problem using deep encoder-decoder architectures. Perhaps the most successful recent method is CycleGAN~\cite{zhu2017unpaired} which uses a deep encoder-decoder architecture (figure~\ref{fig:bias-cyclegan}) together with adversarial and \emph{cycle-consistency} loss terms. The method is demonstrated to succeed and generate high-quality outputs in a variety of tasks such as transforming black and white images to color images, turning horses into zebras and transforming edge images to real (e.g.\ shoes). Many methods followed CycleGAN, modifying the architecture~\cite{liu2017unsupervised, kim2019u} or training objective~\cite{hoshen2018identifying} and improving different aspects of the problem, such as generating more diverse images~\cite{huang2018multimodal, lee2018diverse} and learning from fewer examples~\cite{liu2019few}.

\begin{figure}
    \scriptsize
    \begin{tabular}{cc|ccc|c}
    Input & Target &
    CycleGAN\ddag & MUNIT\ddag & Ours\ddag & Ours\dag \\
    \includegraphics[width=0.117\linewidth]{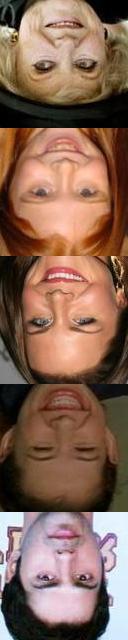} &
    \includegraphics[width=0.117\linewidth]{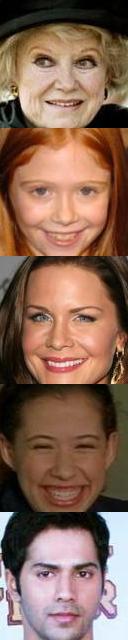} &
    \includegraphics[width=0.117\linewidth]{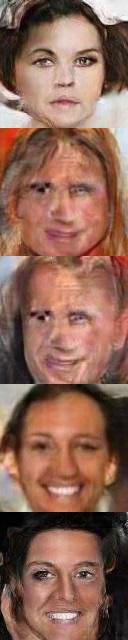} &
    \includegraphics[width=0.117\linewidth]{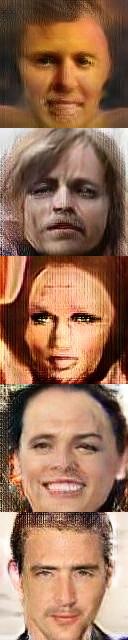} &
    \includegraphics[width=0.117\linewidth]{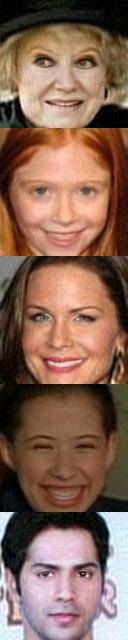} &
    \includegraphics[width=0.117\linewidth]{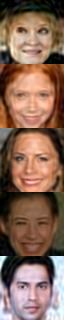} 
    \end{tabular}
    \caption{Deep image-to-image translation methods are biased towards local changes and fail when the true transformation is not local (like \emph{vertical-flip} shown here). Our proposed orthogonal transformation does not suffer from this bias and succeeds in learning non-local transformations.
    \hspace{\textwidth}
    Domains pairing:
    \ddag=Matching pairs exist (shuffled), \dag=Domains contain no matching pairs. }
    \label{fig:nonlocal-fail}
\end{figure}

Although recent unsupervised domain translation (UDT) methods such as CycleGAN and  MUNIT~\cite{huang2018multimodal} have been successful on many unsupervised image-to-image translation problems, it is worth noting that all the problems they are demonstrated on are essentially {\em local}: each pixel in the output image depends only on nearby pixels in the input image and the general image structure is preserved. When this locality assumption does not hold, these methods fail to learn even very simple transformations. Figure~\ref{fig:nonlocal-fail} shows an example. Here the domain $A$ is a set of vertically flipped faces and the domain $B$ is a set of upright faces. All the algorithms have to do is learn to perform a vertical flip. As figure~\ref{fig:nonlocal-fail} shows, CycleGAn and MUNIT fail to learn this very simple transformation. Both methods learn to map an upside-down face to an upright face, but the generate face is distorted and its resemblance to the input face is poor.

Presumably the strong locality bias in modern methods  is due to the large spatial dimension of the encoder-decoder bottleneck that is used in both algorithms (typically $1/4$ of the input resolution) (figure~\ref{fig:bias-cyclegan}). Some of the methods (e.g.\ \cite{zhu2017unpaired}) even have an optional loss term that is simply the $L1$ pixel-to-pixel distance between the input and generated images. Note that several of the UDT papers (e.g.\ \cite{zhu2017unpaired, liu2017unsupervised}) refer to the ill-posedness of the problem. Nevertheless, to the best of our knowledge, all successful methods solve the ill-posedness by using an architecture that is biased towards locality. 

\begin{figure}
    \centering
    \includegraphics[width=\columnwidth]{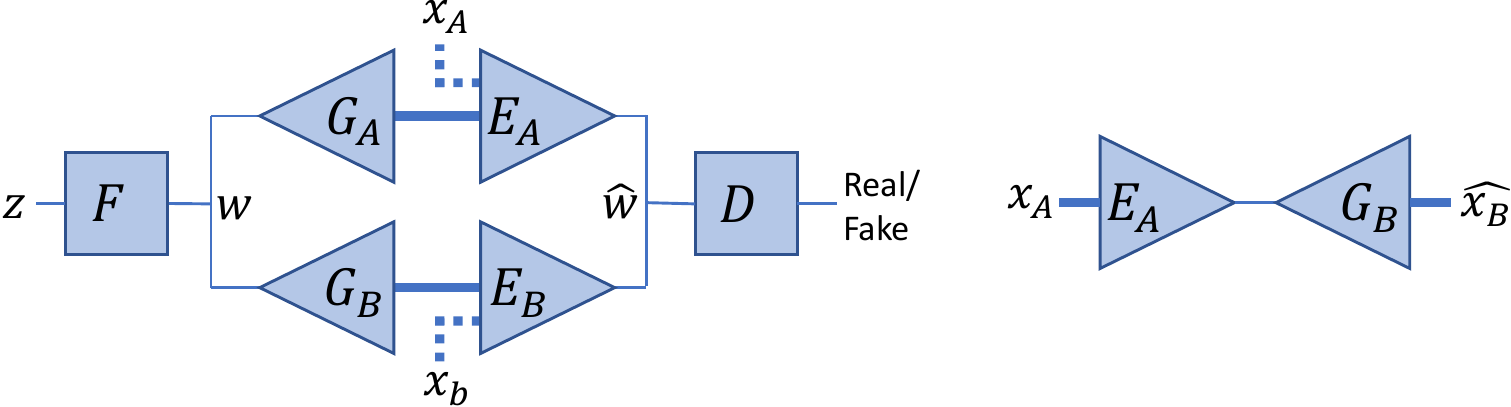}
    \caption{ALAE-based bias-free UDT. Left: To train domain-translation, a second Generator-Encoder pair is added to the base ALAE. The combined loss ensures proper auto-encoding of each domain separately and also cycle-consistency across domains. Right: At inference time, the encoder and generator of the two domains are mixed. Tensors with a spatial dimension are shown as thick lines.}
    \label{fig:unbiased-alae}
\end{figure}

\begin{figure}
    \centering
    \footnotesize
    \begin{tabular}{cccc}
    Input $A$ & $A\rightarrow A$ & $A\rightarrow B$ & $A\rightarrow B\rightarrow A$ \\
    \includegraphics[width=0.15\linewidth]{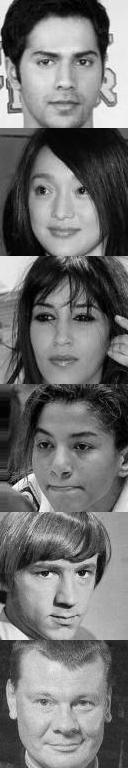} &
    \includegraphics[width=0.15\linewidth]{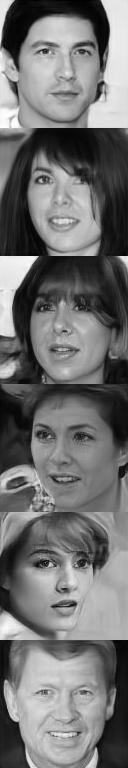} &
    \includegraphics[width=0.15\linewidth]{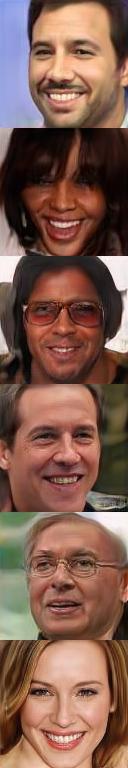} &
    \includegraphics[width=0.15\linewidth]{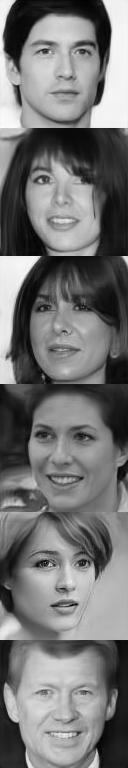} 
    \end{tabular}
    \caption{A deep encoder-decoder architecture without a locality bias (the flat-bottleneck ALAE) converges to an arbitrary solution in the UDT problem ($A\rightarrow B$ has no resemblance to the reconstruction $A\rightarrow A$) even though cycle-consistency is maintained ($A\rightarrow B\rightarrow A$ resembles $A\rightarrow A$).}
    \label{fig:unbiased-fail}
\end{figure}

One way to remove the locality bias is to have an encoder-decoder bottleneck that has \emph{no spatial dimension}. Figure~\ref{fig:unbiased-alae} shows such an architecture based on the very recent ALAE method~\cite{pidhorskyi2020adversarial}, which uses a StyleGAN-based \cite{karras2019style} encoder-decoder \footnote{Simply flattening the bottleneck in the CycleGAN architecture is not possible due to the large number of required parameters. Applying global average-pooling results in strongly distorted outputs images.}. The bottleneck here is a vector of length $512$ but with no spatial dimension and hence there is no particular bias towards local transformations. As shown in figure~\ref{fig:unbiased-fail} when the bias towards local transformations is removed, the method learns an arbitrary mapping between the two domains, even for simple, local transformations. The figure shows the example of colorization. The deep architecture without a locality bias learns to map a gray level face image to a color image \emph{of a different face}, even though cycle-consistency holds.

What is needed therefore is a method that can learn unsupervised image-to-image transformations but without the locality constraint. In this paper we present such a method. It is based on the assumption that the mapping from $A$ to $B$ is {\em a linear, orthogonal transformation}. Although this assumption is clearly restrictive, we show that the method is surprisingly effective -- learning is much easier and faster with these architectures and yet the linear transformations are surprisingly expressive. In particular, we show a number of local problems for which the results of the linear methods are comparable to those of state-of-the-art deep architectures but with a fraction of the training time, and a number of nonlocal problems for which the state-of-the-art fails while linear methods succeed. Code will be made publicly available at \url{https://github.com/eitanrich/lin-im2im}.

\section{Our Approach}

\subsection{Linear Image-to-Image Translation}

We wish to solve the following problem: given a set of images $\mathcal{D}_A$ from domain $A$ and a set of images $\mathcal{D}_B$ from domain $B$ find an {\em orthogonal linear transformation} $T$ that best maps the set $\mathcal{D}_A$ to the set $\mathcal{D}_B$.

\begin{figure}
    \small
    \centering
    \begin{tabular}{cc}
    Inpainting & Colorization\\
    \includegraphics[width=0.42\columnwidth]{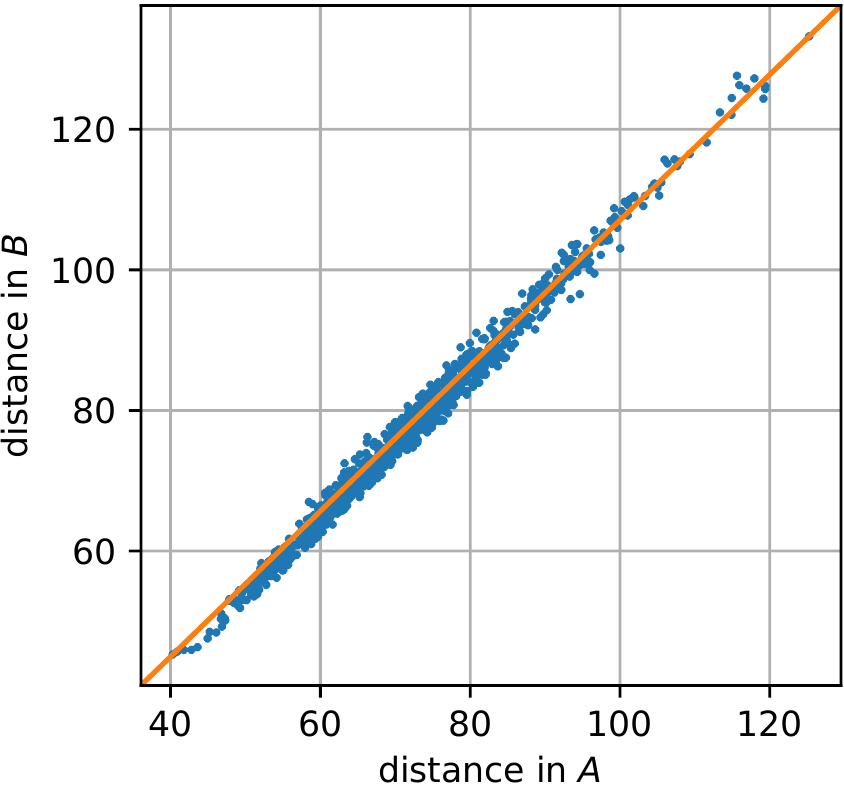} &
    \includegraphics[width=0.42\columnwidth]{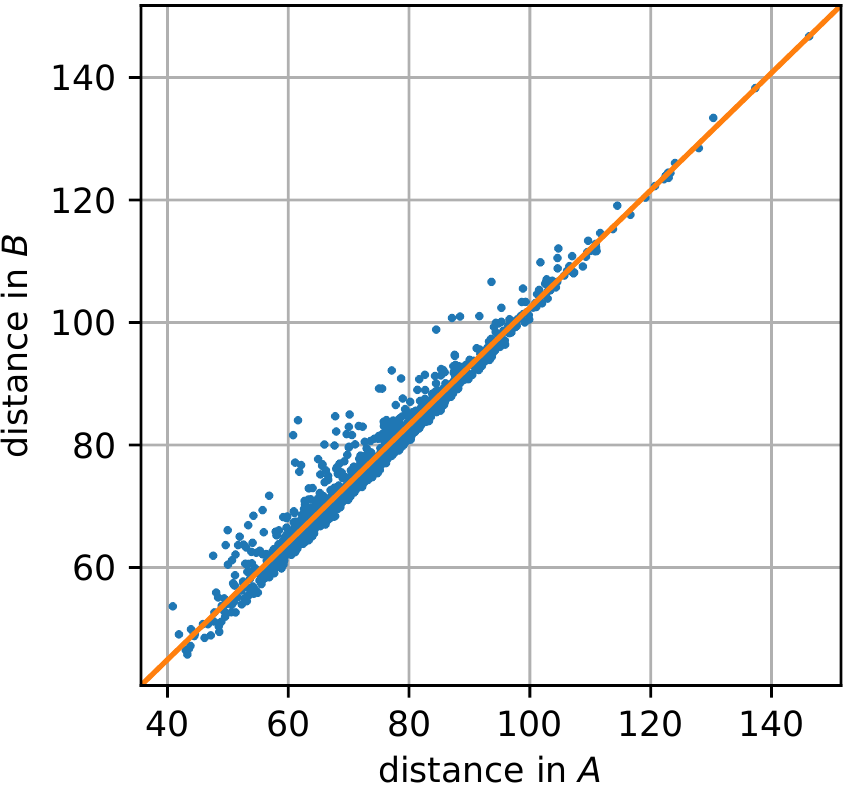} 
    \end{tabular}
    \caption{A scatter plot of the $L2$ distance between two images in domain $A$ and the distance between the corresponding two images in domain $B$ for 1000 randomly selected image pairs from FFHQ. Even though both colorization and inpainting are not invertible transformations, the distances are preserved, suggesting they can be approximated by an orthogonal transformation on a subspace of the original space.  }
    \label{fig:davsdb}
\end{figure}

How restrictive is the  assumption that the transformation is linear and orthogonal? Note that any permutation of the pixels (e.g.\ flipping an image vertically or horizontally) is an orthogonal linear transformation. Less obvious transformations such as inpainting or colorization are also well approximated by an orthogonal transformation. Figure~\ref{fig:davsdb} shows a scatter plot of the distance between two images in domain $A$ and the distance between the corresponding two images in domain $B$ for 1000 randomly selected image pairs. Even though both colorization and inpainting are not invertible transformations, the distances are approximately preserved, suggesting they can be approximated by an orthogonal transformation on a subspace of the original space. Specifically, in the colorizaton example, the gray level images occupy a subspace that is of $1/3$ the dimension of the original RGB images, and yet figure~\ref{fig:davsdb} suggests that if we restrict both RGB and gray level images to lie in a subspace of the same size, then an orthogonal transformation can approximate the mapping.

Even with the restriction to linear orthogonal transformations, the number of such transformations on full images is huge. For a simple example, consider  $128 \times 128$ pixels color images. A linear transformation that maps a set $A$ of such images to a set $B$ of such images can be represented by a matrix of size $49152 \times 49152$ so that the number of free parameters is over 2.4 Billion. How can we learn such a matrix from finite training data? The following observation, shows that we can restrict ourselves to much smaller numbers of free parameters.

{\bf Observation 1:} If every image in $A$ can be approximated as $x_A= W_A z_A$ and each image in $B$ can be approximated as $x_B= W_B z_B$ where $z_A,z_B$ are vectors of length $r$ ($r<d$ the image dimension) and $W_A,W_B$ are orthogonal rectangular matrices (with orthonormal column vectors), then any linear, orthogonal transformation $T$ from $A$ to $B$ can be approximated as:
\begin{equation}
T = W_B Q W_A^T
\label{eq:T-eq}
\end{equation}
where $Q$ is an $r \times r$ orthogonal matrix. \footnote{For simplicity, we assume here that the data has zero mean. In practice we subtract the mean before processing each dataset and add the mean back to produce the final output.}

{\bf Proof:} This follows from the fact that we can write the transformation from $A$ to $B$ ($x_B=T x_A$) as a mapping from $z_A$ to $z_B$ ($z_B= (W_B^T T W_B)z_A$). Defining $Q=W_B^T T W_B$ then it is easy to verify that $Q^T Q = I$.

\subsection{Linear Transformation in PCA Subspace}

Equation~\ref{eq:T-eq} has a simple interpretation as a {\em linear encoder decoder} architecture. The $r \times d$ matrix $W_A^T$ encodes the input image $x_A$ using a vector of length $r$, the matrix $Q$ transforms this vector into the matching encoding of domain $B$, while the $d \times r$ matrix $W_B$ decodes the vector into an image. This is similar to the architecture of modern deep image to image translation methods (e.g.\ figure~\ref{fig:bias-cyclegan}), but with the important difference that the mapping from image to the latent vector is linear. In particular, this means that we can find $W_A,W_B$ easily using Principal Component Analysis (PCA). Orthogonal transformations have been used in the past to model unsupervised domain translation by~\cite{hoshen2018non,xing-etal-2015-normalized, hoshen2018unsupervised}, but they use a deep encoder and decoder rather than the linear one that we use here.

\begin{figure}
    \small
    \centering
    \scriptsize
    \begin{tabular}{cc}
    \includegraphics[width=0.53\linewidth]{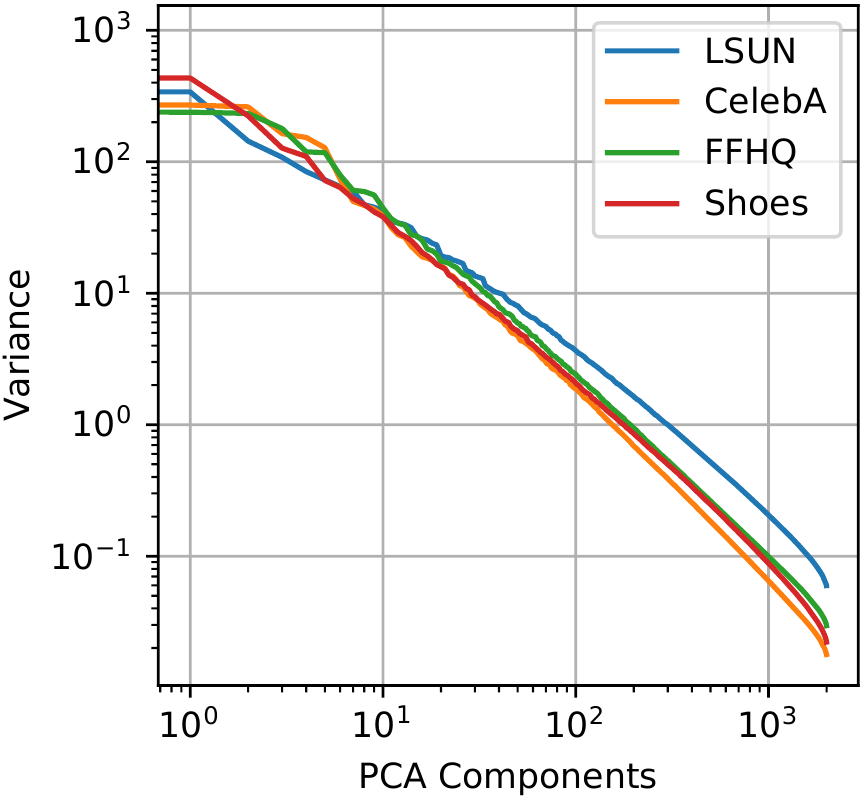} &
    \raisebox{0.5\totalheight}{
    \begin{tabular}{p{2.5mm}cc}
        Orig & \includegraphics[trim=0 0 256px 0, clip, width=0.11\linewidth]{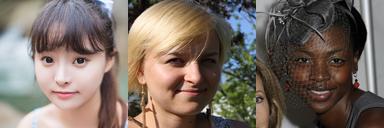} &
        \includegraphics[trim=256px 0 0 0, clip, width=0.11\linewidth]{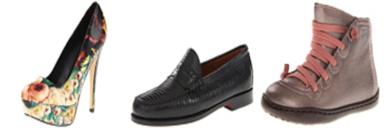} \\
        500 & \includegraphics[trim=0 0 256px 0, clip, width=0.11\linewidth]{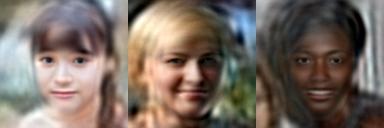} &
        \includegraphics[trim=256px 0 0 0, clip, width=0.11\linewidth]{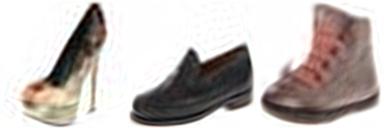} \\
        1000 & \includegraphics[trim=0 0 256px 0, clip, width=0.11\linewidth]{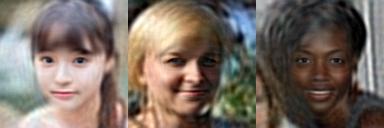} &
        \includegraphics[trim=256px 0 0 0, clip, width=0.11\linewidth]{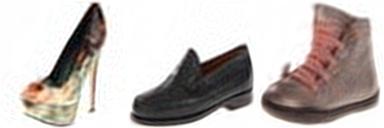} \\
        2000 & \includegraphics[trim=0 0 256px 0, clip, width=0.11\linewidth]{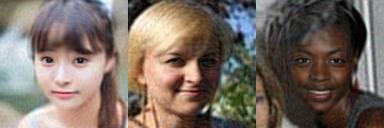} &
        \includegraphics[trim=256px 0 0 0, clip, width=0.11\linewidth]{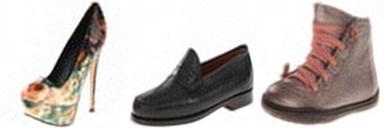}
    \end{tabular}}
    \end{tabular}
    \caption{Left: The PCA spectra of different image datasets behave similarly -- eigenvalues decrease exponentially (with slightly lower rate in the less-structured LSUN dataset). Right: Reconstruction of FFHQ and Shoes test samples ($128\times128$ pixels) using increasing numbers of PCA components.}
    \label{fig:pca}
\end{figure}

 How many PCA coefficients are required? Figure~\ref{fig:pca} shows that the PCA spectrum of commonly used image datasets falls off as a power law (linear in a log-log plot).  This means that images of size $128 \times 128$ pixels can be very well approximated with a {\em linear} encoding and decoding using as few as $2000$ PCA coefficients ($r=2000$). Thus we can reduce our problem to that of finding an orthogonal matrix $Q$ of size $2000 \times 2000$. 

\subsection{Solving for $Q$}
Finding an orthogonal transformation that maps a set of points in $R^n$ to another set of points in $R^n$ is a well-studied problem in computer graphics and computer vision~\cite{besl1992method, myronenko2010point}. It is well known that in the infinite data setting, this problem can be solved efficiently up to a sign ambiguity. We briefly review this solution and then present our algorithm for the specific case of image datasets.

{\bf Observation 2:} Denote by $\Sigma_A,\Sigma_B$ the covariance matrices of the datasets $\mathcal{D}_A=\{x^A_1,\ldots x^A_n\}$, $\mathcal{D}_B=\{x^B_1,\ldots x^B_m\}$. Assume that the true relation between domains $A$ and $B$ is an orthogonal linear transformation $T^*$. If $v$ is an eigenvector of $\Sigma_A$ with eigenvalue $\lambda$ then $T^*v$ is an eigenvector of the covariance $\Sigma_B$ with the same eignevalue. 

Observation 2 implies that if we had infinite data (so that we could reliably estimate $\Sigma_A,\Sigma_B$) and if the eigenvalues of each covariance matrix are unique, then we can recover $Q$ exactly by sorting the eigenvectors in the two domains and mapping the $k$th eigenvector of $\Sigma_A$ to the $k$th eigenvector of $\Sigma_B$. This solution still allows for a {\em sign ambiguity} since if $v$ is an eigenvector of $\Sigma_A$ so is $-v$. Furthermore, for any finite dataset the ordering of the eigenvectors may be changed as a result of sampling different datapoints from each dataset. Therefore the common practice is to initialize an iterative algorithm such as Iterated Closest Points (ICP)~\cite{besl1992method} from multiple initial conditions (each with a different sign chosen for the top eigenvectors)~\cite{hoshen2018non}. 

\begin{figure}
    \centering
    \begin{tabular}{cc}
    \includegraphics[width=0.4\columnwidth]{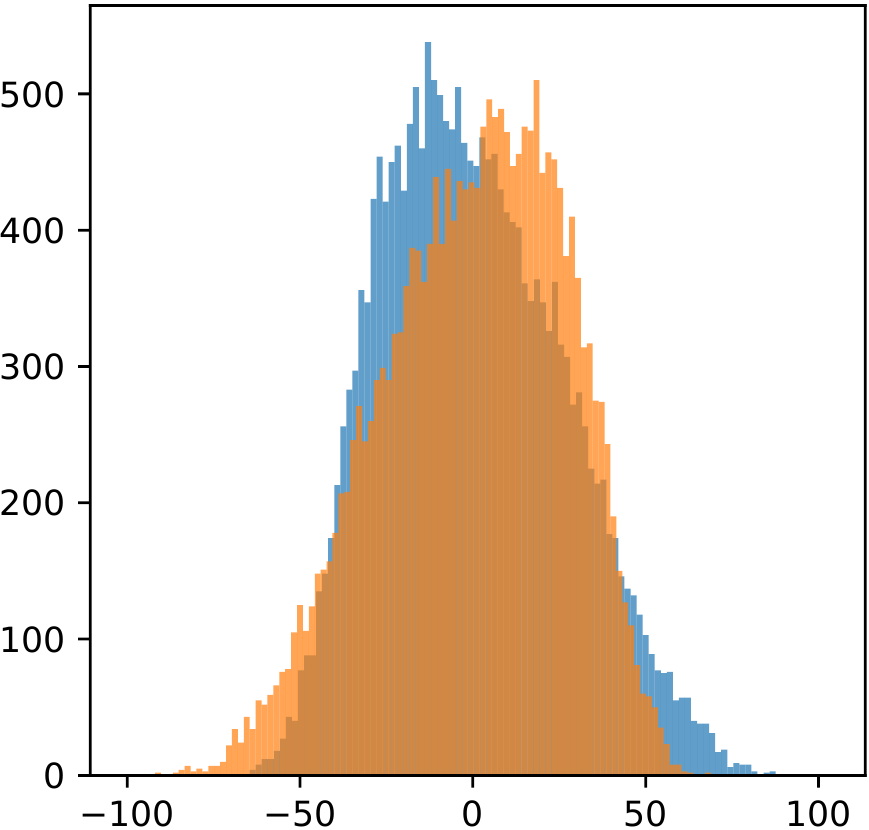} &
    \includegraphics[width=0.4\columnwidth]{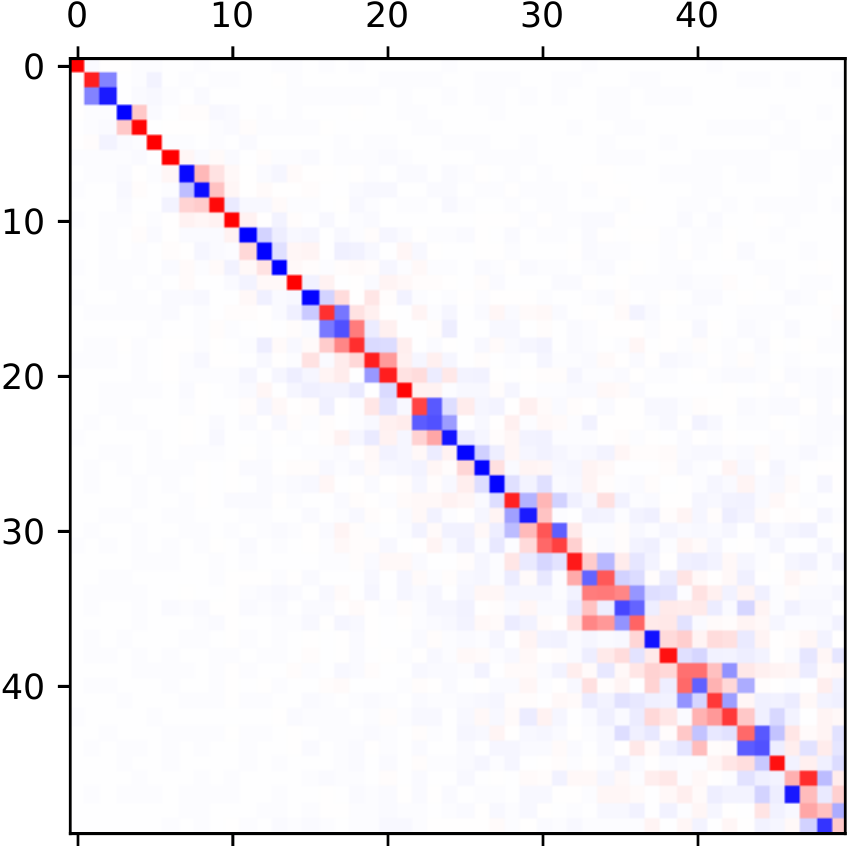} \\
    \includegraphics[width=0.4\columnwidth]{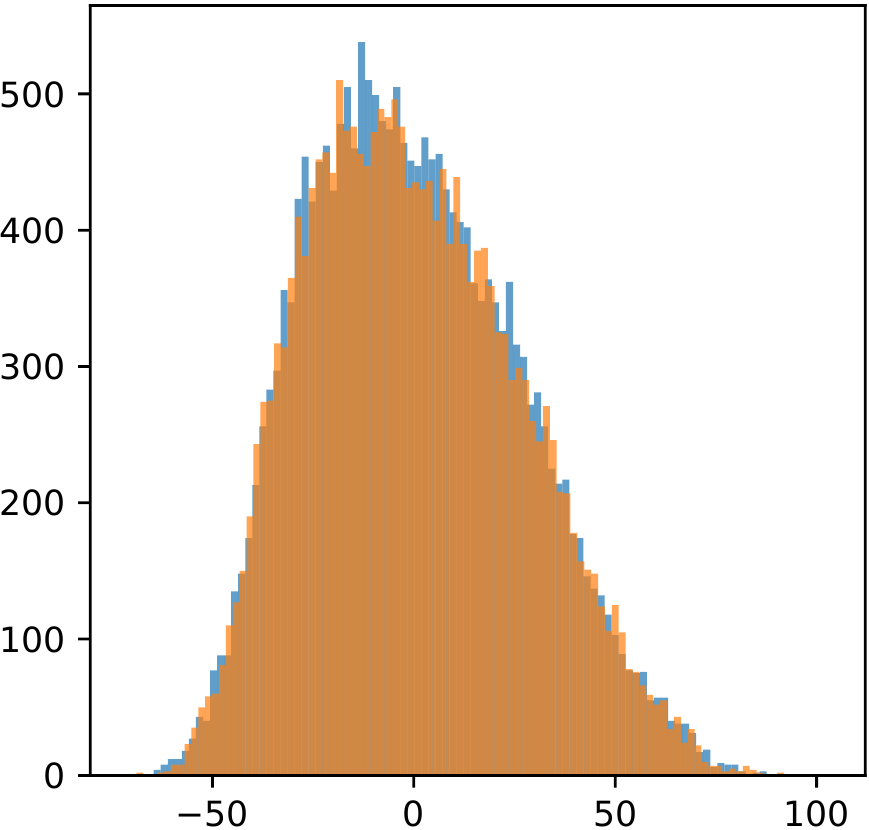} &
    \includegraphics[width=0.4\columnwidth]{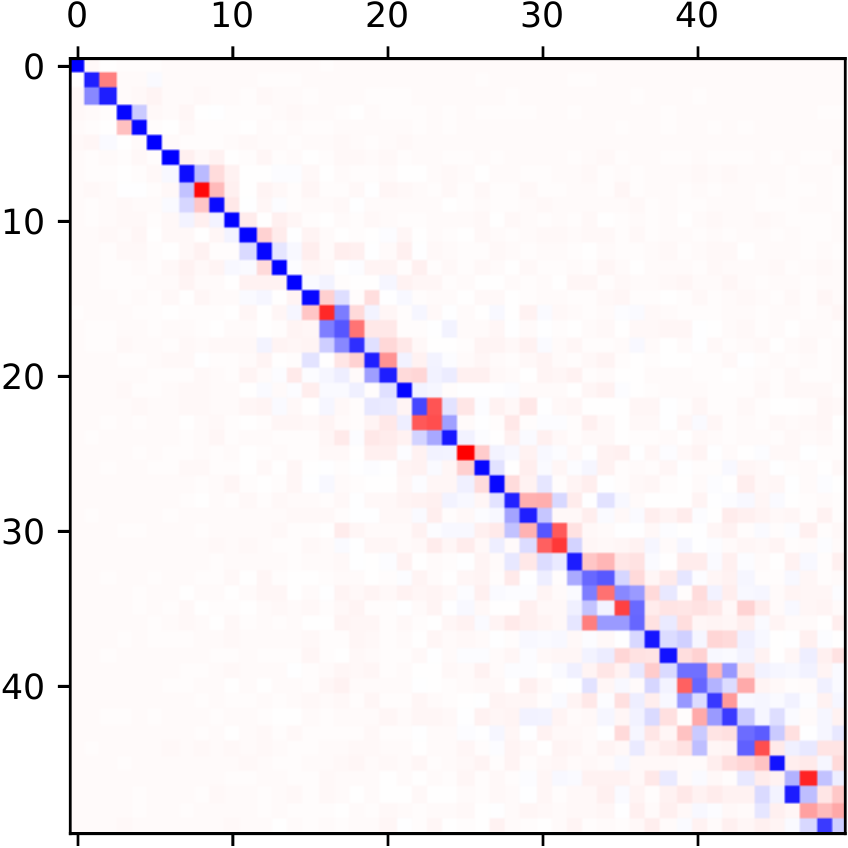} \\
    \end{tabular}
    \caption{Using the skewness to resolve PCA sign ambiguity. Left: Histograms of $W^A_0 X_A$ and $W^B_0 X_B$ (first coordinate in the data PCA embedding) before aligning the skewness (top) and after aligning (bottom). Right: The top-left block (first 50 coordinates) of $Q$ before and after aligning the skewness (red=negative). Aligning the skewness makes $Q$ much closer to $I$.}
    \label{fig:skew-and-q}
\end{figure}

We make two modifications to the standard approach. First, we have found that for image datasets, the sign ambiguity can be mostly resolved using the \emph{skewness} of the distribution (see figure~\ref{fig:skew-and-q}). Specifically, given an eigenvector $v$ we calculate the distribution of $v^T x$ on the dataset and calculate the skewness of that distribution:
\[
skew(v)= \sum_i (v^T x_i - \mu)^3
\]
If the skewness is negative, we replace $v$ with $-v$. Second, in order to improve the accuracy of ICP, we use the ``best buddy'' heuristic~\cite{dekel2015best}: if $x_B$ is the closest point to $T x_A$ we also require that $x_A$ be the closest point $T^T x_B$. See algorithm~\ref{algo:icp} for details. Once a set of corresponding pairs was found (e.g.\ using ICP), the orthogonal transformation $Q$ is estimated using the \emph{Procrustes} method (lines $12-13$). 

Note that our algorithm can be applied even when the two sets do not contain matching pairs. The existence of matching pairs improves the ICP convergence speed and the data efficiency (in the nonmatching case, the size of the training sets needs to be sufficiently large so that cross-domain nearest-neighbors will be reasonably similar). 

\begin{algorithm}
 \DontPrintSemicolon
 \small
 \KwIn{$\mathcal{D}_A=\{x^A_1,\ldots x^A_n\}$, $\mathcal{D}_B=\{x^B_1,\ldots x^B_m\}$, $r$}
 \KwResult{Orthogonal transformation $T:A\rightarrow B$}
 Compute $W_A, W_B$: $r$ principal components of $\mathcal{D}_A, \mathcal{D}_B$\;
 Fix eigenvectors sign for positive skew\;
 Compute PCA embedding $\{z^A_1,\ldots z^A_n\}$, $\{z^B_1,\ldots z^B_m\}$\;
 Initialize $Q \gets I$\;
 \While{not converged}{
  $A \gets \emptyset$, $B \gets \emptyset$\; 
  \For{$i \gets 1$ \textbf{to} $n$}{
  $j \gets \argmin_{j'}||z^A_i Q - z^B_{j'}||$\;
  $k \gets \argmin_{k'}||z^A_{k'} Q - z^B_j||$\;
  \If{$k=i$}{
  $A\text{.insert-row}(z^A_i)$, $B\text{.insert-row}(z^B_j)$\;
  }
  }
  $U, S, V \gets \operatorname{SVD}(A^T B)$\;
  $Q \gets UV$\;
 }
 \Return{$T \gets W_A^T Q W_B$}\;
 \caption{Orthogonal UDT in PCA subspace}
 \label{algo:icp}
\end{algorithm}

In practice we find this algorithm to be very fast (training times of a few seconds for commonly used datasets, whereas training CycleGAN can take more than a day).

\section{Results}

\subsection{Datasets and Transformations}

We conduct our experiments using two popular face datasets -- CelebA \cite{liu2015faceattributes} and FFHQ \cite{karras2019style} and one dataset of non-face images -- Shoes \cite{yu2014fine}. We trained all models on images resized to $128 \times 128$ pixels. To train our method we used $20$K images from each domain. For the SOTA deep methods we used $50$K images from CelebA and the entire $60$K FFHQ train images. All results are shown on (unseen) test images.

We tested the following synthetic transformations:
\begin{center}
\begin{tabular}{|l|l|l|} \hline
    \small
     Task Name & Domain A & Domain B \\
    \hline
     \emph{vflip} &  Vertically flipped & Original\\
     \emph{rotate} &  90 degrees rotated left& Original\\
     \emph{colorize} & Grayscale image & Original\\
     \emph{inpaint} & Set to 0 in a mask & Original\\
     \emph{edges-to-real} & Sobel edges & Original\\
     \emph{super-res} & Reduced size by 8 & Original\\
    \hline
\end{tabular}
\end{center}

In most tests, sets $A$ and $B$ contained \emph{matching} images -- set $A$ was generated by applying the selected transformation to the original images and then rearranging them in random order (shuffling). We also tested the \emph{nonmatching} case in which the original dataset was first randomly split in half and then the selected transformation was applied to one half. 

\subsection{Qualitative Results}

\begin{figure*}
    \small
    \centering
    \begin{tabular}{c|cc|cc|cc|cc}
        Target & \multicolumn{2}{c|}{Rotate} & \multicolumn{2}{c|}{Colorize} & \multicolumn{2}{c}{Super-resolution $\times8$} & \multicolumn{2}{c}{Inpaint} \\
        \includegraphics[width=0.08\linewidth]{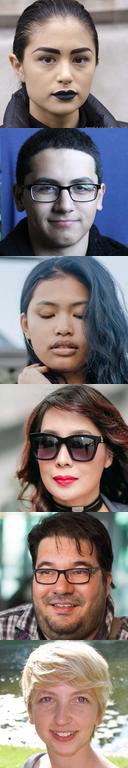} &
        \includegraphics[width=0.08\linewidth]{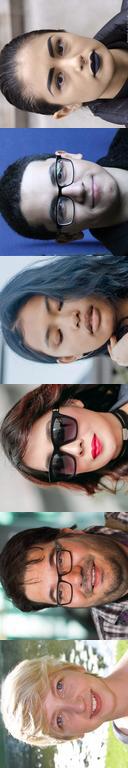} &
        \includegraphics[width=0.08\linewidth]{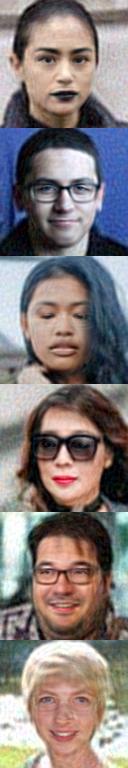} &
        \includegraphics[width=0.08\linewidth]{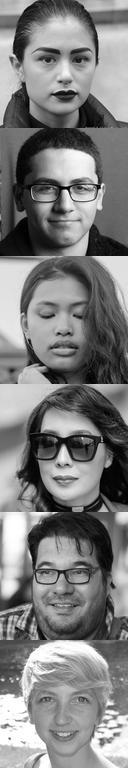} &
        \includegraphics[width=0.08\linewidth]{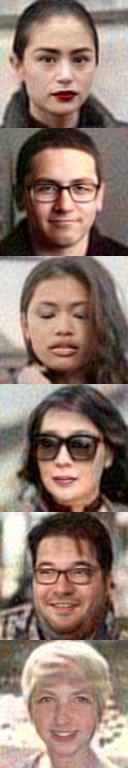} &
        \includegraphics[width=0.08\linewidth]{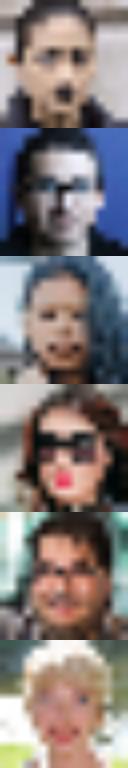} &
        \includegraphics[width=0.08\linewidth]{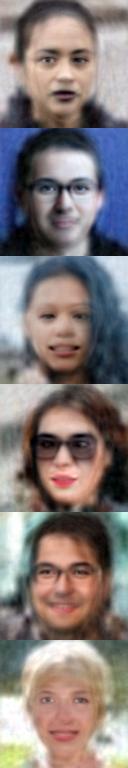} &
        \includegraphics[width=0.08\linewidth]{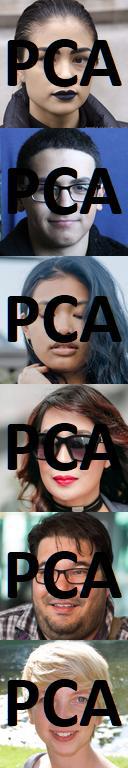} &
        \includegraphics[width=0.08\linewidth]{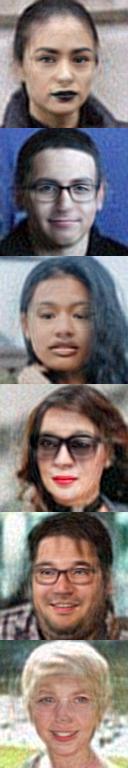}
    \end{tabular}
    \caption{Domain translation using our proposed orthogonal transformation in PCA domain, demonstrated on different FFHQ tasks. Training on 20K samples of each domain with unpaired matching pairs. Results shown on unseen test images.}
    \label{fig:res-ffhq-matching}
\end{figure*}

\begin{figure}
    \small
    \centering
    \begin{tabular}{cc|cc|cc}
        \multicolumn{2}{c|}{Rotate} & \multicolumn{2}{c|}{Colorize} & \multicolumn{2}{c}{Inpaint} \\
        \includegraphics[width=0.12\linewidth]{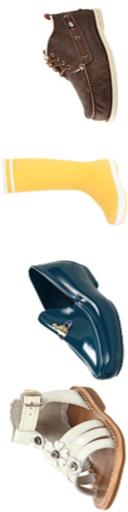} &
        \includegraphics[width=0.12\linewidth]{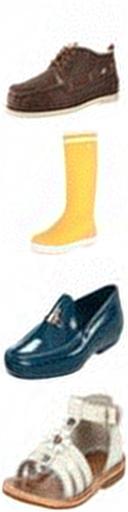} &
        \includegraphics[width=0.12\linewidth]{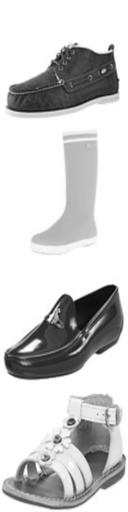} &
        \includegraphics[width=0.12\linewidth]{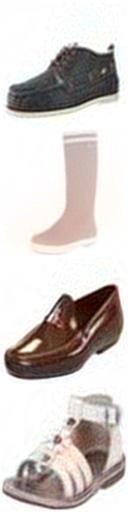} &
        \includegraphics[width=0.12\linewidth]{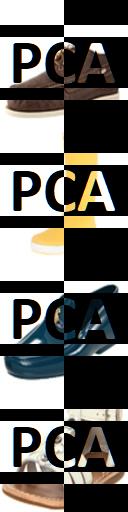} &
        \includegraphics[width=0.12\linewidth]{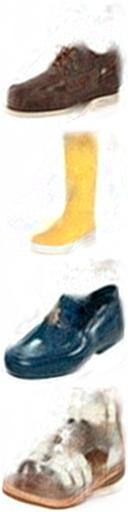} 
    \end{tabular}
    \caption{Results of our paired PCA-based linear transformation on non-face images -- Shoes.}
    \label{fig:res-shoes}
\end{figure}

Figure~\ref{fig:res-ffhq-matching} shows the  orthogonal transformation learned by our proposed method (algorithm~\ref{algo:icp}) on several transformations applied to the FFHQ dataset. As can be seen, our method can learn a variety of relations between $A$ and $B$ and the learned transformation can be applied successfully to unseen test images. Similar results are shown in figure~\ref{fig:res-shoes} for the \emph{Shoes} dataset. Results for CelebA are shown in figure~\ref{fig:nonlocal-fail}.

Figure~\ref{fig:res-ffhq-paired} shows the learned transformation in the \emph{paired} setting for more challenging tasks -- converting edge images to real images and in-painting where a large part of the original image is hidden. For the paired (supervised) case, the correspondence finding step (lines $6-11$ of algorithm~\ref{algo:icp}) is not required and the transformation is found in a single iteration using the two paired sets.

\subsection{Quantitative Evaluation}

Since we use synthetic true transformations, we can measure the quality of the learned transformation by comparing the transformed images to the ``ground truth'' target images.
We used two commonly used image similarity measures -- the mean squared error (MSE) and the structural similarity (SSIM) \cite{wang2004image}.
Table~\ref{tab:eval-quality} lists the transformation quality achieved by our method compared to the SOTA deep neural-network methods, as well as the training times. We compare both the \emph{local} case, which is suitable to the architectural bias in the deep methods and the \emph{nonlocal} case. For \emph{colorization}, which is a \emph{local} transformation, our method achieves similar results to CycleGAN and MUNIT, but at less than $1/1000$ of the training time. For the two \emph{nonlocal} tasks, our method achieves significantly better results than the deep methods.

\begin{table*}
    \centering
    \begin{threeparttable}
    \caption{Quantitative evaluation of unsupervised image-to-image translation methods}
    \centering
    \begin{tabular}{|c|ccc|ccc|ccc|ccc|}
        \hline
        Task & \multicolumn{3}{c|}{CycleGAN\ddag} & \multicolumn{3}{c|}{MUNIT\ddag} & \multicolumn{3}{c|}{Ours\ddag}& \multicolumn{3}{c|}{Ours\dag} \\
        & MSE & SSIM & $T$[h] &  MSE & SSIM & $T$[h] &  MSE & SSIM & $T$[h] &  MSE & SSIM & $T$[h] \\
        \hline
        CelebA-colorize &
        0.0066 & {\bf 0.914} & 49 &
        0.0256 & 0.750 & 52 &
        {\bf 0.0043} & 0.883 & {\bf 0.04} &
        0.0071 & 0.761 & 0.04 \\
        \hline
        CelebA-vflip &
        0.1167 & 0.358 & 43 &
        0.1084 & 0.333 & 48 &
        {\bf 0.0012} & {\bf 0.917} & {\bf 0.04} &
        0.0041 & 0.780 & 0.04\\
        \hline
        FFHQ-rot90 &
        0.1267 & 0.302 & 39 &
        0.1220 & 0.268 & 39 &
        {\bf 0.0023} & {\bf 0.870} & {\bf 0.05} &
        0.0335 & 0.381 & 0.05\\
        \hline
    \end{tabular}
    \begin{tablenotes}
    \item MSE is the mean-squared error between the input and target images,  
    SSIM~\cite{wang2004image} estimates the perceptual similarity (higher the better) and $T$ is the total training time in hours (including data loading).
    \item Domains pairing:
    \ddag=Matching pairs exist (shuffled), 
    \dag=No matching pairs.
    \end{tablenotes}
    \label{tab:eval-quality}
    \end{threeparttable}
\end{table*}

\section{Limitations and Possible Extensions}
\begin{figure}
    \small
    \centering
    \begin{tabular}{cc|cc}
        \multicolumn{2}{c|}{Edges-to-Real} & \multicolumn{2}{c}{Inpaint} \\
        \includegraphics[width=0.14\linewidth]{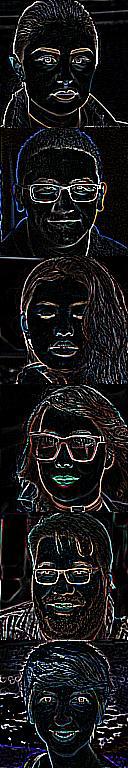} &
        \includegraphics[width=0.14\linewidth]{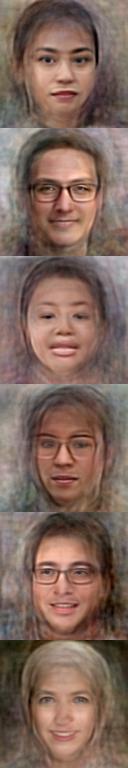} &
        \includegraphics[width=0.14\linewidth]{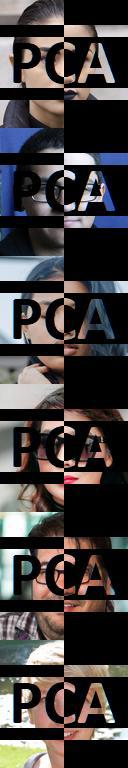} &
        \includegraphics[width=0.14\linewidth]{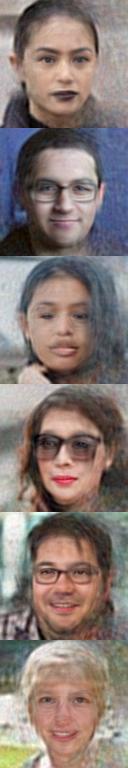} 
    \end{tabular}
    \caption{Our proposed linear translation handling more challenging tasks in the supervised (paired) setting. Note that for the \emph{edges-to-real} task, an unrestricted linear translation was chosen instead of an orthogonal one.}
    \label{fig:res-ffhq-paired}
\end{figure}

Figure~\ref{fig:res-ffhq-paired} shows some limitations of our method. 
By definition, the main limitation of the proposed linear transformation is modeling true transformations that are very \emph{non-linear}. An example is \emph{edges to real-images}, in which our results are inferior to those of deep encoder-decoder methods that can model non-linear transformations. For such a setting, even a fully supervised application of our method (when the pairing between $x_A$ and $x_B$ are given to the algorithm) does not give good results, even if we allow arbitrary linear transformations.

Another limitation is that very complex image domains may require a large number of PCA coefficients to represent properly and still, very fine details may not be well reconstructed when passing through the low-dimensional PCA subspace.

\begin{figure}
    \centering
    \footnotesize
    \begin{tabular}{c|c|cc|c}
    Input $A$ & Lin. $A\rightarrow B$ & $A\rightarrow A$ & $A\rightarrow B$ & $A\rightarrow B\rightarrow A$ \\
    \includegraphics[width=0.14\linewidth]{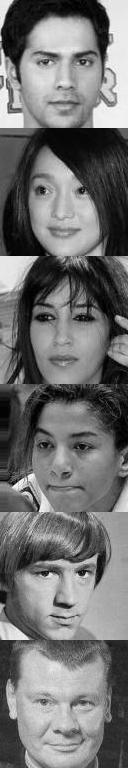} &
    \includegraphics[width=0.14\linewidth]{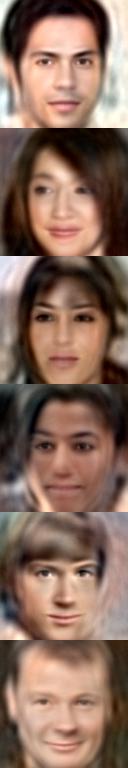} &
    \includegraphics[width=0.14\linewidth]{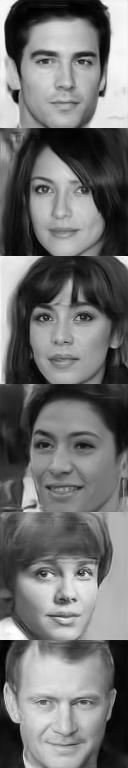} &
    \includegraphics[width=0.14\linewidth]{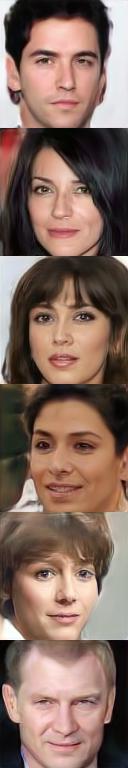} &
    \includegraphics[width=0.14\linewidth]{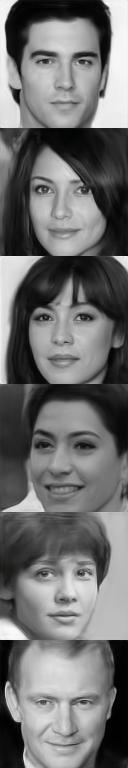} 
    \end{tabular}
    \caption{ALAE (similar to figure~\ref{fig:unbiased-fail}, but with linear orthogonal transformation as a regularization term). Now the transformation $A\rightarrow B$ is very similar to the reconstruction $A\rightarrow A$, but with the added color information required for domain $B$ -- the transformation is no longer arbitrary.}
    \label{fig:unbiased-plus-orthogonal}
\end{figure}

A possible solution (figure~\ref{fig:unbiased-plus-orthogonal}) to the above limitations is combining the orthogonal transformation, which is easy to compute and not limited to local-changes with a bias-free deep encoder-decoder architecture, which excels at generating realistic and sharp images. The orthogonal transformation can guide the convergence of the over-expressive deep architecture towards the desirable solution. This approach can be thought of as a generalization of the ``identity $L1$'' loss term used by CycleGAN and other methods, replacing the naive assumption that $T(x_A)\approx x_A$ (e.g.\ zebras look like horses) with the more general assumption that the relation between $A$ and $B$ is approximately linear. Considering figure~\ref{fig:unbiased-alae}, the suggested loss term is therefore: $\mathcal{L}_{orthogonal}=||G_A(w)W_A^TQ-G_B(w)W_B^T||_2$, where $W_A, W_B$ and $Q$ are pre-computed using algorithm~\ref{algo:icp}. As can be seen in figure~\ref{fig:unbiased-plus-orthogonal}, the deep encoder-decoder learns a $A\rightarrow B$ transformation that is more accurate than the linear transformation used for regularization (we used only $300$ PCA coefficients in this test, making the linear orthogonal transformation blurry). Note that the deep encoder-decoder reconstruction itself (Input $A$ vs $A\rightarrow A$) is not perfect. This is a current limitation of ALAE and other StyleGAN based auto-encoders.

\section{Discussion}

Many recent deep encoder-decoder based methods demonstrate success on the inherently ill-posed unsupervised image-to-image translation problem. These methods employ sophisticated training methods such as adversarial and cycle-consistency loss. Despite their success, however, we show that they rely heavily on a particular locality bias that is embodied in the architectures they use -- assuming the local image structure is preserved between the two domains. We show that these methods fail when the true transformation is nonlocal and when the architecture bias is removed.

As an alternative, we presented a very different bias -- \emph{orthogonal linear transformations}. We show that such transformations can approximate a wide range of true domain relationships and solve different tasks such as in-painting and colorization, in addition to any geometric transformation. We suggest a highly efficient algorithm that expresses the orthogonal transformation in PCA subspace, requiring only a small fraction of the number of parameters compared to linear transformation in the full image space. The learning time for the linear transformations is a few seconds, compared to GPU-days for the deep methods.

We are not suggesting that linear methods should replace the deep encoder-decoder methods, however, we believe that presenting the effectiveness and surprising versatility of the orthogonal transformations can be of benefit to image-to-image research community and can help expanding the range of solved problems beyond local transformations.

\bibliographystyle{IEEEtran}
\bibliography{IEEEabrv,refs.bib}

\begin{thebibliography}{10}
\providecommand{\url}[1]{#1}
\csname url@samestyle\endcsname
\providecommand{\newblock}{\relax}
\providecommand{\bibinfo}[2]{#2}
\providecommand{\BIBentrySTDinterwordspacing}{\spaceskip=0pt\relax}
\providecommand{\BIBentryALTinterwordstretchfactor}{4}
\providecommand{\BIBentryALTinterwordspacing}{\spaceskip=\fontdimen2\font plus
\BIBentryALTinterwordstretchfactor\fontdimen3\font minus
  \fontdimen4\font\relax}
\providecommand{\BIBforeignlanguage}[2]{{%
\expandafter\ifx\csname l@#1\endcsname\relax
\typeout{** WARNING: IEEEtran.bst: No hyphenation pattern has been}%
\typeout{** loaded for the language `#1'. Using the pattern for}%
\typeout{** the default language instead.}%
\else
\language=\csname l@#1\endcsname
\fi
#2}}
\providecommand{\BIBdecl}{\relax}
\BIBdecl

\bibitem{zhu2017unpaired}
J.-Y. Zhu, T.~Park, P.~Isola, and A.~A. Efros, ``Unpaired image-to-image
  translation using cycle-consistent adversarial networks,'' in
  \emph{Proceedings of the IEEE international conference on computer vision},
  2017, pp. 2223--2232.

\bibitem{liu2017unsupervised}
M.-Y. Liu, T.~Breuel, and J.~Kautz, ``Unsupervised image-to-image translation
  networks,'' in \emph{Advances in neural information processing systems},
  2017, pp. 700--708.

\bibitem{kim2019u}
J.~Kim, M.~Kim, H.~Kang, and K.~Lee, ``U-gat-it: unsupervised generative
  attentional networks with adaptive layer-instance normalization for
  image-to-image translation,'' \emph{arXiv preprint arXiv:1907.10830}, 2019.

\bibitem{hoshen2018identifying}
Y.~Hoshen and L.~Wolf, ``Identifying analogies across domains,'' in
  \emph{International Conference on Learning Representations}, 2018.

\bibitem{huang2018multimodal}
X.~Huang, M.-Y. Liu, S.~Belongie, and J.~Kautz, ``Multimodal unsupervised
  image-to-image translation,'' in \emph{Proceedings of the European Conference
  on Computer Vision (ECCV)}, 2018, pp. 172--189.

\bibitem{lee2018diverse}
H.-Y. Lee, H.-Y. Tseng, J.-B. Huang, M.~Singh, and M.-H. Yang, ``Diverse
  image-to-image translation via disentangled representations,'' in
  \emph{Proceedings of the European conference on computer vision (ECCV)},
  2018, pp. 35--51.

\bibitem{liu2019few}
M.-Y. Liu, X.~Huang, A.~Mallya, T.~Karras, T.~Aila, J.~Lehtinen, and J.~Kautz,
  ``Few-shot unsupervised image-to-image translation,'' in \emph{Proceedings of
  the IEEE International Conference on Computer Vision}, 2019, pp.
  10\,551--10\,560.

\bibitem{pidhorskyi2020adversarial}
S.~Pidhorskyi, D.~A. Adjeroh, and G.~Doretto, ``Adversarial latent
  autoencoders,'' in \emph{Proceedings of the IEEE/CVF Conference on Computer
  Vision and Pattern Recognition}, 2020, pp. 14\,104--14\,113.

\bibitem{karras2019style}
T.~Karras, S.~Laine, and T.~Aila, ``A style-based generator architecture for
  generative adversarial networks,'' in \emph{Proceedings of the IEEE
  conference on computer vision and pattern recognition}, 2019, pp. 4401--4410.

\bibitem{hoshen2018non}
Y.~Hoshen and L.~Wolf, ``Non-adversarial unsupervised word translation,''
  \emph{arXiv preprint arXiv:1801.06126}, 2018.

\bibitem{xing-etal-2015-normalized}
\BIBentryALTinterwordspacing
C.~Xing, D.~Wang, C.~Liu, and Y.~Lin, ``Normalized word embedding and
  orthogonal transform for bilingual word translation,'' in \emph{Proceedings
  of the 2015 Conference of the North {A}merican Chapter of the Association for
  Computational Linguistics: Human Language Technologies}.\hskip 1em plus 0.5em
  minus 0.4em\relax Denver, Colorado: Association for Computational
  Linguistics, May{--}Jun. 2015, pp. 1006--1011. [Online]. Available:
  \url{https://www.aclweb.org/anthology/N15-1104}
\BIBentrySTDinterwordspacing

\bibitem{hoshen2018unsupervised}
Y.~Hoshen and L.~Wolf, ``Unsupervised correlation analysis,'' in
  \emph{Proceedings of the IEEE Conference on Computer Vision and Pattern
  Recognition}, 2018, pp. 3319--3328.

\bibitem{besl1992method}
P.~J. Besl and N.~D. McKay, ``Method for registration of 3-d shapes,'' in
  \emph{Sensor fusion IV: control paradigms and data structures}, vol.
  1611.\hskip 1em plus 0.5em minus 0.4em\relax International Society for Optics
  and Photonics, 1992, pp. 586--606.

\bibitem{myronenko2010point}
A.~Myronenko and X.~Song, ``Point set registration: Coherent point drift,''
  \emph{IEEE transactions on pattern analysis and machine intelligence},
  vol.~32, no.~12, pp. 2262--2275, 2010.

\bibitem{dekel2015best}
T.~Dekel, S.~Oron, M.~Rubinstein, S.~Avidan, and W.~T. Freeman, ``Best-buddies
  similarity for robust template matching,'' in \emph{Proceedings of the IEEE
  conference on computer vision and pattern recognition}, 2015, pp. 2021--2029.

\bibitem{liu2015faceattributes}
Z.~Liu, P.~Luo, X.~Wang, and X.~Tang, ``Deep learning face attributes in the
  wild,'' in \emph{Proceedings of International Conference on Computer Vision
  (ICCV)}, 2015.

\bibitem{yu2014fine}
A.~Yu and K.~Grauman, ``Fine-grained visual comparisons with local learning,''
  in \emph{Proceedings of the IEEE Conference on Computer Vision and Pattern
  Recognition}, 2014, pp. 192--199.

\bibitem{wang2004image}
Z.~Wang, A.~C. Bovik, H.~R. Sheikh, and E.~P. Simoncelli, ``Image quality
  assessment: from error visibility to structural similarity,'' \emph{IEEE
  transactions on image processing}, vol.~13, no.~4, pp. 600--612, 2004.

\end{thebibliography}

\end{document}